# A Tensor-Based Sub-Mode Coordinate Algorithm for Stock Prediction

Jieyun Huang
Key Laboratory of Trustworthy
Distributed Computing and
Service, Ministry of Education
Beijing University of Posts and
Telecommunications
Beijing , China
hjy_@bupt.edu.cn

Yunjia Zhang
Key Laboratory of Trustworthy
Distributed Computing and
Service, Ministry of Education
Beijing University of Posts and
Telecommunications
Beijing , China
2011213120@bupt.edu.cn

Jialai Zhang
School of Electronic
Information
Sichuan University
Chengdu, China
1010271376@qq.com

Xi Zhang
Key Laboratory of Trustworthy
Distributed Computing and
Service, Ministry of Education
Beijing University of Posts and
Telecommunications
Beijing , China
zhangx@bupt.edu.cn

*Abstract*— The investment on the stock market is prone to be affected by the Internet. For the purpose of improving the prediction accuracy, we propose a multi-task stock prediction model that not only considers the stock correlations but also supports multi-source data fusion.
  Our proposed model first utilizes tensor to integrate the multi-sourced data, including financial Web news, investors' sentiments extracted from the social network and some quantitative data on stocks. In this way, the intrinsic relationships among different information sources can be captured, and meanwhile, multi-sourced information can be complemented to solve the data sparsity problem. Secondly, we propose an improved sub-mode coordinate algorithm (SMC). SMC is based on the stock similarity, aiming to reduce the variance of their subspace in each dimension produced by the tensor decomposition. The algorithm is able to improve the quality of the input features, and thus improves the prediction accuracy. And the paper utilizes the Long Short-Term Memory (LSTM) neural network model to predict the stock fluctuation trends. Finally, the experiments on 78 A-share stocks in CSI 100 and thirteen popular HK stocks in the year 2015 and 2016 are conducted. The results demonstrate the improvement on the prediction accuracy and the effectiveness of the proposed model.

## I. INTRODUCTION

Stock market are strongly affected by various sources of information. New information related to markets may change the opinions of investors and then affect stock price. Essentially, stock information is multifaceted and interrelated. Generally, stock-related information can be roughly categorized into the quantitative data and the qualitative descriptions of the financial standings. Quantitative analysis aims to study economic and business data to predict stock price movement. Nevertheless, quantitative data cannot entirely convey the boundless variety of financial standings of firms [1]. Qualitative information lies in the textual descriptions from various data sources including the web media and social media, which is actually complementary to quantitative data.

Investigating the joint effect of multiple information sources on stock movements remains a challenge. A common strategy in previous studies has been to concatenate the features of multiple information sources into one compound feature vector. However, as the dimension of the vector increases, solution strategies are subject to the "curse of dimensionality" [2]. Besides, the contextual co-occurrence relations among the various information sources are reduced or possibly eliminated. To address the challenge, Li Q et al. presented a Global Dimensionality-Reduction (GDR) algorithm that captures the static and dynamic interconnections among the various information sources [3]. However, each stock prediction in that algorithm is modeled as an individual task and thus learned independently, regardless of the correlations among the stocks.

In this paper, we present a novel tensor-based computational framework that can effectively predict stock price movements by fusing various sources of information. We propose to combine the stock-related events extracted from the Web news, the users' sentiments on social media and some quantitative data, and capture the intrinsic relationships among different information sources. In addition, an improved sub-mode coordinate algorithm (SMC) is presented. SMC is based on the stock similarity, aiming to reduce the variance of their subspace in each dimension produced by the tensor decomposition. The algorithm is able to improve the quality of the input features, and thus enhances the prediction accuracy. And the paper utilizes the Long Short-Term Memory (LSTM) neural network model to predict the stock fluctuation trends. The model is evaluated on 78 A-share stocks in CSI 100 and thirteen popular HK stocks in the year 2015 and 2016. The results demonstrate the improvement on the prediction accuracy and the effectiveness of the proposed model.
  In the remainder of this article, Section 2 introduces the related work. We give the preliminary in Section 3. Section 4 describes the system framework. In Section 5, evaluation on real data shows the effectiveness of the proposed approach. Finally, we conclude the paper in Section 6.

## II. RELATED WORKS

A company's stock price reflects the value of the future profits of the company. Once a related news or policy is reported, it will have a certain impact on the stock market. In addition, the release of news will also have a certain impact on investors' sentiments and investment decisions, and indirectly on the stock market. Various studies have found that financial news has dramatic effects on the stock price [4, 5, 6] and some researches show that the stock market is event-driven [7, 8]. Structured events have been extracted as tuples including the agent, predicate and object from news documents in [9]. Deep learning-based method is proposed to learn event representations to capture syntactic and semantic information based on word embedding [10]. There are also a series of researches trying to apply sentiment analysis on information sources to analyze the impacts of sentiments on stock market volatility. A hybrid approach is introduced in [11] for stock sentiment analysis based on news articles on companies. News sentiments are measured and the combined effect of Web news and social media on stock markets are studied in [12].

Recent studies investigate the joint impacts of events and the sentiments. Li Q et al. explore the impact of stock investors' sentiment and news events on the stock market [13]. There is a mutual relationship between news events and stock investors' sentiment. Therefore, the use of multi-dimensional data to jointly predict the stock market volatility is reasonable. Prior studies concatenate the data of these different factors into one vector. However, there are hidden internal links between different dimensions of data. In 2015, Li Q et al. proposed a stock price prediction model based on tensor regression in [14]. Compared with the previous method for linearly concatenating heterogeneous data features, the forecasting capability was significantly improved. Because the tensor structure can capture the potential relationship between heterogeneous data, we use the tensor fusion method to fuse multi-source data. Li Q et al. proposed a tensor-based GDR algorithm in [3] to obtain the tensor after dimensionality reduction for stock price regression. The sub-mode coordinate algorithm proposed in this paper is improved on the basis of it, and the tensor-decomposed sub-planes are coordinated and updated by the association relationship between stocks.

In the field of deep learning, the Recurrent neural network is widely used to tackle the problem of stock price prediction [15]. The Long Short-Term Memory (LSTM) is a specific type of recurrent neural network which overcomes some of the problems of recurrent networks [16]. An LSTM memory cell stores a value, for either long or short time periods. This is achieved by using an identity activation function for the memory cell. In this way, when an LSTM network is trained with backpropagation through time, the gradient does not tend to vanish. Our research investigates whether the time series model can help to predict the stock price movement.

III. PRELIMINARIES

In this part, we briefly introduce the mathematical notations and the tensor operations used in this paper. Tensors are high-order arrays that generalize the notions of vectors and matrices. In this paper, we use a third-order tensor, representing a 3-dimensional array. Scalars are 0th-order tensors and denoted by lowercase letters, e.g., a. Vectors are 1st-order tensor and denoted by boldface lowercase letters, e.g., **a**. Matrices are 2nd-order tensor and denoted by boldface capital letters, e.g., ***X***, and 3rd-order tensors are denoted by calligraphic letters, e.g., ***A***. The $i$th entry of a vector **a** is denoted by $a_i$, element $(i,j)$ of a matrix ***X*** is denoted by $x_{ij}$, and element $(i,j,k)$ of a 3rd-order tensor $A$ is denoted by $a_{ijk}$. The $i$th row and the $j$th column of a matrix ***X*** are denoted by $x_{i:}$ and $x_{:j}$, respectively. Alternatively, the $i$th row of a matrix, $\boldsymbol{a}_{i:}$, can also be denoted as $\boldsymbol{a}_i$.

The norm of a tensor $\mathcal{X} \in \mathbb{R}^{I \times J \times K}$ is defined as

$$\|\mathcal{X}\|_F = \sqrt{\sum_{i_1}^{I_1} \sum_{i_2}^{I_2} \cdots \sum_{i_N}^{I_N} \mathcal{X}(i_1, i_2, \ldots, i_n)^2}$$

This is analogous to the matrix *Frobenius norm*, which is denoted as $\|\mathcal{X}\|$ for a matrix ***X***. The outer product of several vectors can get a tensor. For instance, as Figure 1 shows that the third-order tensor $X$ can be expressed as $\mathcal{X} = \mathbf{a} \circ \mathbf{b} \circ \mathbf{c}$, that is the outer product operation of vectors $\mathbf{a}, \mathbf{b}, \mathbf{c}$.

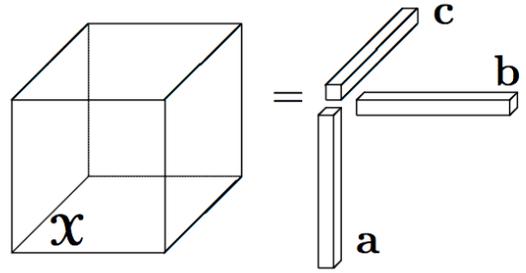

Fig. 1. Tensor uses outer product representation.

The *n-mode product* of a tensor $C \in R^{I_1 \times I_2 \times I_3}$ with a matrix $U \in R^{I_n \times J}$ denoted by $C \times_n U$ is a tensor of size $I_1 \times \cdots \times I_{n-1} \times J \times I_{n+1} \times \cdots \times I_N$ with the elements $(C \times_n U)_{i_1,\ldots,i_{n-1},r,i_{n+1},\ldots,i_n} = \sum_{i_n=1}^{I_n} a_{i1,i2,iN} u_{inj}$.

The *Tucker factorization* of a tensor $X \in \mathbb{R}^{I \times J \times K}$ is defined as:
$$X = C \times_1 U \times_2 V \times_3 W$$
Here, $U \in \mathbb{R}^{I \times R_1}$, $V \in \mathbb{R}^{J \times R_2}$ and $W \in \mathbb{R}^{K \times R_3}$ are the factor matrices and can be thought of as the principal components in each mode. The tensor $C \in \mathbb{R}^{R_1 \times R_2 \times R_3}$ is the core tensor and its entries show the level of interaction between the different components. The reconstructed tensor is derived by multiplying the core tensor and the three factor matrices. It can be observed that tensor decomposition and reconstruction have updated the value for each existing entries indicating its importance and filled some new entries showing the latent relationships. Generally, tensor factorization can be regarded as an extension of the matrix decomposition. During the process of decomposition, data can be projected in the subspaces, which includes latent significance. Figure 2 presents the result of the *Tucker factorization*.

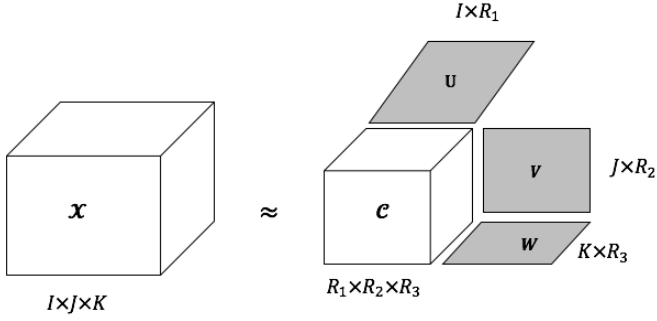

Fig. 2. The Tucker factorization of a tensor

## IV. THE PROPOSED METHOD

In this section, we depict the proposed system framework for stock price movement prediction. Specifically, the proposed sub-mode coordinate algorithm(SMC) which is used to modify the quality of the input tensor will be introduced in detail.

### A. The System Framework

We employ the historical stock quantitative data, Web news articles and social media data to construct a third-order tensor jointly to predict the stock price movements. The overall system framework is shown in Fig. 3, which is comprised of four major parts: 1) The stock quantitative data, the events and sentiments extracted from news articles and social media to build the stock movement tensor; 2) The stock correlation matrix construction with the multi-sourced data; 3) Enhance a sub-mode coordinate algorithm (SMC) to capture the dynamic correlation among a sequence of tensors for on stock and a set of tensors among different stocks at the same time; 4) The reconstructed tensor by the SMC algorithm would be the input of the time series neural network to train the model and finally make stock predictions. We will describe the first two parts in this section, and the last two parts will be introduced in the next section.

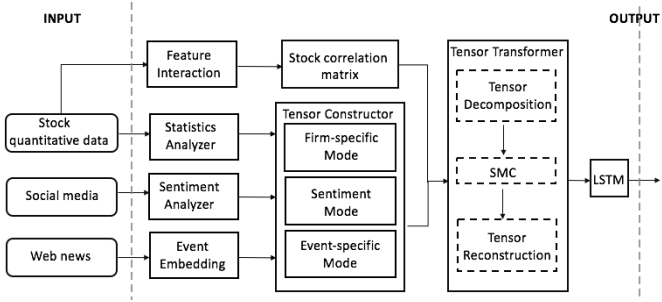

Fig. 3. The system framework.

### B. The Sub-mode Coordinate Algorithm(SMC)

The proposed sub-mode coordinate algorithm(SMC) is an approach to reducing the dimension of the tensor. The SMC algorithm considers both the similarity among multiple stocks and the stock similarity among different time. In the horizontal aspect, there exists similar stock movement trend at different time for a stock. Vertically, there also exists the similar movement trend for different stocks at the same time. The similar stock fluctuation means the similar tensor in the general geometric structure. Therefore, the SMC algorithm aims to make the decomposed corresponding sub-spaces of the similar tensors close to each other as much as possible. And then obtain the similar reconstructed tensor after the modification of the SMC algorithm. Generally speaking, the SMC algorithm can be considered as a method for modifying the initial input tensor for extracting the more critical features and then improving the quality of the input. The overall SMC algorithm framework is shown in Fig. 4.

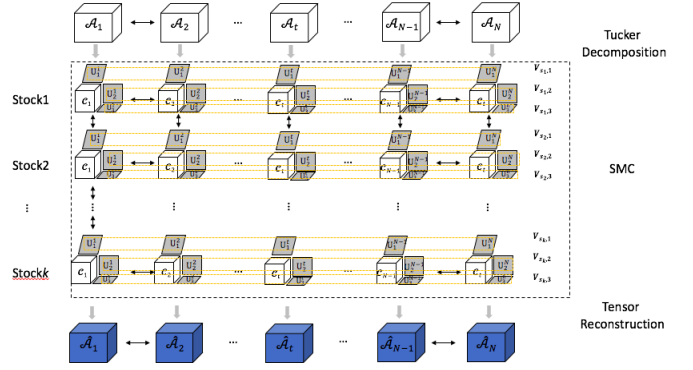

Fig. 4. The SMC algorithm framework.

Firstly, establish a sequenced set of tensors for every stock $s$ at each day $t$ $\{\mathcal{A}_{s,1}, \mathcal{A}_{s,2}, \ldots, \mathcal{A}_{s,t}\}$ as the input of SMC. After decomposing the tensor $\mathcal{A}_{s,t} \in \mathbb{R}^{I_1 \times I_2 \times I_3}$ with Tucker, three sub-spaces on each dimension and a core tensor with smaller dimension will be obtained $\mathcal{A}_{s,t} = \mathcal{C}_{s,t} \times_1 U_1^{s,t} \times_2 U_2^{s,t} \times_3 U_3^{s,t}$. The core tensor $\mathcal{C}_{s,t} \in \mathbb{R}^{D_1 \times D_2 \times D_3}$ contains the intrinsic relationship among three dimensions. And $U_k^{s,t} \in \mathbb{R}^{I_k \times D_k}$ is the sub-space on the $k$-th dimension that is obtained by decomposing the initial tensor. For the tensors with similar stock movement, the SMC algorithm aims to reconstruct those initial decomposed sub-spaces with the modification matrices $V_k \in \mathbb{R}^{I_k \times J_k}$, where $J_k \leq I_k$ and make the reconstructed sub-spaces similar to each other as much as possible. Therefore, the objective function is depicted as follow:

$$\min_{V_k} \mathcal{J}(V_k) = \sum_{s=1}^{S} \sum_{i=1}^{T} \sum_{j=i}^{T} \left\| V_k^T U_k^{s,i} - V_k^T U_k^{s,j} \right\|^2 w_{s,i,j} + \sum_{t=1}^{T} \sum_{s=1}^{S} \sum_{m=s}^{S} \left\| V_k^T U_k^{m,t} - V_k^T U_k^{s,t} \right\|^2 z_{t,s,m}$$

$w_{s,i,j}|_{i=1,i>j}^{T}$ denotes the similarity weight of stock $s$ at the time $i$ and $j$. If the variation of p-change between the time $i$ and $j$ is less than the similarity threshold $\mathcal{E}_1$, the tensor can be considered as similar in general and set the similarity weight $w_{s,i,j}$ as 1. The similarity matrix $W_s$ is an upper triangular matrix, where $w_{s,i,j}$ equals to zero if $i$ is less than or equal to $j$. The similarity matrix $W_s$ can be represented as follow:

$$w_{s,i,j} = \begin{cases} 1, & i \leq j \text{ and } |y_{s,i} - y_{s,j}|/y_{s,j} \leq E_1 \\ 0 & \end{cases}$$

$z_{t,s,m}$ denotes the similarity weight between the stock $s$ and stock $m$ at the time $t$. The stock correlation matrix $Z_t$ is obtained by the feature interaction method which is proposed in [17].

We select the ADAM optimization algorithm to optimize the objective function using the tensorflow framework. The ADAM algorithm dynamically adjusts the learning rate based on the moment estimation and the torque estimation of the gradient of each parameter. ADAM is also based on the gradient descent method, but the learning step of each iteration has a definite range. It will not cause a large learning step due to a large gradient, and the value of the parameter is relatively stable. The training results in the modification matrix $V_k$ ($k = 1, 2, 3$) for different dimension.

The general process of SMC is described as Algorithm 1.

---

ALGORITHM I. THE SUB-MODE COORDINATE ALGORITHM (SMC)

Input: Tensors $\{\mathcal{A}_{s,1}, \mathcal{A}_{s,2}, ..., \mathcal{A}_{s,t}\}$, where ($s$=1,2, …, S), the similarity matrix $W_s$ and the stock correlation matrix $Z_t$, the corresponding similarity thresholds $\mathcal{E}_1$, $\mathcal{E}_2$, the maximum iteration times $IterMax$.

Output: The modification matrices $V_k \in \mathbb{R}^{I_k \times J_k}$, where ($k$=1, 2,3).

1. Tucker decomposition $A_{s,t} = C_{s,t} \times_1 U_1^{s,t} \times_2 U_2^{s,t} \times_3 U_3^{s,t}$
2. Initialize the modification matrices $V_{s,k} \in \mathbb{R}^{I_k \times J_k}$, and set the random numbers among 0 and 1 in each entry.
3. Initialize the moment vector $m_0 \leftarrow 0$ and the torque vector $v_0 \leftarrow 0$, the iteration times $it$=0.
4. Set the step $\alpha = 0.001$, the decay rate of the moment estimation $\beta_1$, the decay rate of the torque estimation $\beta_2$.
5. while ($V_{s,k}^{it}$ not converge or $it$<$IterMax$) do
6.    for each $s$ do
7.       while ($i$<=T) do
8.          while ($j$<=T) do
9.             $Loss^{it+1} = Loss^{it} + \|V_k^T U_k^{s,i} - V_k^T U_k^{s,j}\|^2 w_{s,i,j}$
10.    end for
11.    for each $t$ do
12.       while ($s$<=S) do
13.          while ($m$<=S) do
14.             $Loss^{it+1} = Loss^{it} + \|V_k^T U_k^{m,t} - V_k^T U_k^{s,t}\|^2 z_{t,s,m}$
15.    end for
16.    $it$=$it$+1
17.    $grad_{s,k}^{it} \leftarrow \nabla_{V_{s,k}^{it}} Loss^{it-1}$
18.    $m_{it} \leftarrow \beta_1 \cdot m_{it-1} + (1-\beta_1) \cdot grad_{s,k}^{it}$
19.    $v_{it} \leftarrow \beta_2 \cdot v_{it-1} + (1-\beta_2) \cdot {grad_{s,k}^{it}}^2$
20.    $\hat{m}_{it} \leftarrow m_{it}/(1-\beta_1)$
21.    $\hat{v}_{it} \leftarrow v_{it}/(1-\beta_2)$
22.    $V_{s,k}^{it} \leftarrow V_{s,k}^{it-1} - \alpha \cdot \hat{m}_{it}/\hat{v}_{it}$
23. end while
24. return $V_{s,k}(k = 1,2,3)$

---

## V. EXPERIMENTS

### A. Data Collection and Description

Our proposed method is evaluated on two datasets, including the China A-share stock market data and HK stock market data during Jan. 1, 2015 and Dec. 31, 2016, and the multi-sourced data involves:

*a) Quantitative data:* The quantitative data of stocks of the two datasets are both collected from *Wind* which contains various financial data. We select the stock turnovers, P/E ratios, P/B ratios and PCF ratios that are commonly used indices for stock trading and evaluation as quantitative indexes. In addition, we select industry index to calculate the stock correlation matrix.

*b) Web news data:* We collect 90,361 news articles for A-share stocks and HK stocks respectively from Wind, including the titles and the publication time. Each article is assigned to the corresponding stock. These Web news are originally aggregated by Wind from major financial news websites in China, such as *http://finance.sina.com.cn* and *http://www.hexun.com.* These titles are then processed to extract events as described in our previous work. The data is publicly available on [18].

*c) Social media data:* The sentiments for A-share stocks are extracted from Guba. Guba is an active financial social media where investors can post their opinions and each stock has its own discussion site. Totally, we collect 1179,926 postings from Jan.1, 2015 to Dec. 31, 2016 for the stocks in the experiment. For each posting, it includes the content, user ID, the title, the number of comments and clicks, and the publication time are extracted. We also have made this dataset publicly available [19].

In the experiments, we use the data in the first 9 months as the training set and the last 3 months as the testing set. And 47.1% of all the samples present upward trend, while 52.3% present downward trend and 0.6% keep still. To remove the obstacles and obtain deterministic price moving trends, we set the movement scope threshold as 2%. In particular, when the price change ratio of a stock is larger than 2% (or smaller than -2%) on a trading day, its price movement direction is considered as up (or down). Otherwise, this sample is considered as little fluctuation (or still) and excluded from our experiments.

### B. Baselines and Evaluation metrics

*a) SVM:* Directly concatenate the stock quantitative features, event mode features and sentiment mode features as a linear vector, and use them as the input of SVM for prediction.

*b) PCA+SVM:* The Principle Component Analysis (PCA) technique is applied to reduce the dimensions of the original concatenated vector, and then the new vector is used as the input of SVM.

*c) TeSIA:* The tensor-based learning approach proposed in [14] is the state-of-art baseline which utilizes multi-source information. Specially, it uses a third-order tensor to model the firm-mode, event-mode and sentiment-mode data. The method constructs an independent tensor for every stock on each trading day, without considering the correlations between stocks.

*d) CMT:* CMT is the model of our previous work, which contains two auxiliary matrices and a tensor are factorized together. The default CMT is using the couple stock correlation method.

*e) GDR+TeSIA:* GDR is a tensor dimensionality reduction algorithm which is presented in [3]. We apply it before *TeSIA* regression.

*f) SMC+TeSIA:* The model takes SMC as a substitute for GDR in the previous model e, and meanwhile, still uses TeSIA to predict the stock movements. Therefore, we are able to prove the effectiveness of this model in contrast with model e.

*C. Prediction Results*

Following the previous studies [20], the standard measures of accuracy (ACC) and Matthews Correlation Coefficient (MCC) are used as evaluation metrics. Larger values of the two metrics mean better performance.

Table 1 shows the average stock movement prediction accuracy for all the CSI 100 stocks and 13 HK market stocks during the testing period. We choose the model that directly concatenate the stock quantitative features, event mode features and sentiment mode features as a linear vector to prove the effectiveness of the tensor-based multi-source data fusion model. Besides, the different tensor dimensionality reduction algorithm is used to compare with the improved SMC algorithm. Furthermore, choosing the model that takes stock's prediction as regression learning to prove the effectiveness of the time series model.

According to the experimental results of the two methods for SVM and SVM+PCA in Table 1, the prediction accuracy of the method by directly concatenating multi-source data into linear vectors is only between 55% and 58%. For the five models TeSIA, GDR+TeSIA, SMC+TeSIA and SMC+LSTM, which use tensor-based data fusion, the prediction accuracy rate of the models reaches more than 60%. Compared to the linear concatenating method, the performance of the TeSIA which is the tensor-based data fusion method also increases by 3%-5%. Therefore, it is very meaningful to use tensor to fuse multi-source heterogeneous data, and the performance of model prediction is greatly improved. In addition, through the experimental results of TeSIA, GDR+TeSIA, and SMC+TeSIA, the tensor regression method using the improved SMC algorithm has a prediction accuracy rate of 62.29% over the CSI100 dataset, which is 0.83% more accurate than the tensor regression method with GDR dimensionality reduction and increases by nearly 0.1 on the MCC. Similarly, the accuracy of the SMC+TeSIA method on Hong Kong stocks is also significantly higher than that of GDR+TeSIA, with an increase of 1.13%. Therefore, it is shown that the improved sub-mode coordinate algorithm proposed in this paper has a significant improvement effect on the prediction of stock movements. In the end, the prediction accuracy rate obtained by using the SMC+LSTM time series model is about 1% higher than that of SMC+TeSIA based on the tensor regression method. Therefore, it indicates that the stock trend forecasting model is based on certain rules in time series, so the use of the time series model has a certain significance for stock forecasting.

Table 1: Results of different stock prediction model

| Method | CSI100 | | HK | |
|---|---|---|---|---|
| | ACC | MCC | ACC | MCC |
| **SVM** | 55.37% | 0.014 | 55.13% | 0.08 |
| **SVM+PCA** | 57.50% | 0.104 | 56.07% | 0.092 |
| **TeSIA** | 60.63% | 0.190 | 60.38% | 0.205 |
| **CMT** | 62.50% | 0.409 | 61.73% | 0.331 |
| **GDR+TeSIA** | 61.46% | 0.203 | 60.92% | 0.201 |
| **SMC+TeSIA** | 62.29% | 0.302 | 62.05% | 0.24 |
| **SMC+LSTM** | 62.97% | 0.371 | 63.31% | 0.398 |

VI. CONCLUSION

In this section, we depict the proposed system framework for stock price movement prediction. We propose to combine the stock-related events extracted from the Web news, the users' sentiments on social media and some quantitative data, which capture the intrinsic relationships among different information sources. In the previous work, only the degree of similarity between the same stocks at different times in the horizontal direction was taken into account for adjustment of the tensor factorization adjustment. In this work, the SMC algorithm is proposed that also considered the correlations between different stocks at the same time in the vertical direction, and use it to produce a certain degree of synergy. Therefore, the SMC algorithm uses the similarity of the factorized matrices from a more comprehensive way to make adjustments between the sub-planes, which is equivalent to the normalization work of modifying the features of the input samples. The tensor-based long and short-term memory time-series model is used to capture the regular relationship between stocks and to predict the stock fluctuations. Finally, experiments show that our method can effectively improve the accuracy of prediction.

ACKNOWLEDGMENT

This work was supported in part by the National Key Research and Development Program of China (No. 2016QY03D0605, No, 2017YFB0803301), the Natural Science Foundation of China (No. 61300014, 61372191, 61472263), the Project on the Integration of Industry, Education and Research of Guangdong Province (No. 2016B090921001), and DongGuan Innovative Research Team Program (No. 201636000100038).